\documentclass{article}
\usepackage[letterpaper, top=1in, bottom=1in, left=1in, right=1in]{geometry}
\usepackage[utf8]{inputenc}
\usepackage{enumitem}
\usepackage{tabularx}
\usepackage{booktabs} 
\usepackage{threeparttable}
\usepackage{float}

\usepackage{graphicx} 
\usepackage{hyperref}
\usepackage{enumitem}
\usepackage[numbers]{natbib}
\usepackage{csquotes}
\usepackage{comment}
\usepackage{lipsum}
\usepackage[dvipsnames]{xcolor}
\usepackage[font=bf,skip=\baselineskip]{caption}
\usepackage[toc,page]{appendix}
\usepackage{listings}
\newcommand\surl[1]{{\footnotesize \url{#1}}}

\makeatletter
\def\verbatim@font{\linespread{1}\normalfont\ttfamily}
\makeatother

\definecolor{codegreen}{rgb}{0,0.6,0}
\definecolor{codegray}{rgb}{0.5,0.5,0.5}
\definecolor{codepurple}{rgb}{0.58,0,0.82}
\definecolor{backcolour}{rgb}{0.95,0.95,0.92}

\lstdefinestyle{mystyle}{
    backgroundcolor=\color{backcolour},   
    commentstyle=\color{codegreen},
    keywordstyle=\color{magenta},
    numberstyle=\tiny\color{codegray},
    stringstyle=\color{codepurple},
    basicstyle=\ttfamily\footnotesize,
    breakatwhitespace=false,         
    breaklines=true,                 
    captionpos=b,                    
    keepspaces=true,                 
    numbers=left,                    
    numbersep=5pt,                  
    showspaces=false,                
    showstringspaces=false,
    showtabs=false,                  
    tabsize=2
}

\usepackage{multirow}  
\usepackage{placeins}
\usepackage{abstract}
\usepackage{siunitx}

\usepackage{caption}    
\usepackage{subcaption} 
\usepackage{authblk}

\newcommand{\bottomstrut}[1]{\rule[-#1]{0pt}{#1}}

\title{You've Changed: Detecting Modification of Black-Box Large Language Models}
\author{Alden Dima, James Foulds, Shimei Pan, Philip Feldman}
\affil{University of Maryland, Baltimore County\\ \texttt{\{dimaa1, jfoulds, shimei, feld1\}@umbc.edu}}
\date{\today}

\begin{document}

\maketitle

\begin{abstract}

Large Language Models (LLMs) are often provided as a service via an API, making it challenging for developers to detect changes in their behavior. We present an approach to monitor LLMs for changes by comparing the distributions of linguistic and psycholinguistic features of generated text. Our method uses a statistical test to determine whether the distributions of features from two samples of text are equivalent, allowing developers to identify when an LLM has changed. We demonstrate the effectiveness of our approach using five OpenAI completion models and Meta's Llama 3 70B chat model. Our results show that simple text features coupled with a statistical test can distinguish between language models. We also explore the use of our approach to detect prompt injection attacks. Our work enables frequent LLM change monitoring and avoids computationally expensive benchmark evaluations. 

\end{abstract}
\section{Introduction}

The rise of Large Language Models (LLM) has brought a new paradigm for Natural Language Processing (NLP), dubbed ``pre-train, prompt, and predict'' which replaces more traditional approaches where an existing language model is fine tuned for new tasks~\cite{Liu2023-ep}. In essence, the prompt allows for the transformation of a problem at hand into one that the language model has been pre-trained on so that the answer can be predicted by filling in the appropriate slots in the prompt~\cite{Liu2023-ep}.
As a result, LLMs are increasingly being used as components in software development where the combination of an LLM plus a prompt is used to implement functionality, particularly in applications that require NLP~\cite{Ma2023-jk}.

However, this newfound flexibility has shifted developer focus to LLMs, which are often provided as ``black boxes'' hidden behind an API~\cite{Ma2023-jk}.  A recent post in social media drew significant attention when a user sought confirmation that GPT-4 appeared to be getting ``dumber.''~\cite{Carter2023-hy}
Chen et al. evaluated GPT-3.5 and GPT-4 across multiple tasks and found that their performance and behavior changed significantly between two releases spaced several months apart~\cite{Chen2023-ib}.
For certain tasks, LLM performance improved, while for others it worsened.

LLM updates force downstream developers to constantly adapt their products as these often silent updates can change LLM performance and behavior~\cite{Ma2023-jk}.
These updates can include model upgrades to a newer version, or replacing the model with a variant having a different number of parameters or quantization.
They may even include subtle changes to the model weights via techniques such as the use of control vectors.
One such impact is ``prompt brittleness'' which occurs because prompts are often tuned for a particular LLM and are sensitive to subtle changes in the language model~\cite{Ma2023-jk}.
A silent update can require a rewriting the prompts used to implement certain functions.
In the case of GPT-3.5 and GPT-4, some have suggested that developers find alternatives until the performance fluctuations are addressed~\cite{Ortiz2023-ev}

Regression testing is typically used to ensure that software changes do not introduce unexpected behaviors in previously implemented and tested functionality~\cite{Beizer1990-bp} but the stochastic nature of LLMs complicates regression testing of LLM-based applications~\cite{Ma2023-jk}.
Developers must account for considerable nondeterministic behavior and be prepared to address large sample sizes in their tests~\cite{Ma2023-jk}.
For example, code generation applications can experience significant nondeterminism in their generated outputs. Ouyang et al. report that the fraction of coding task results with no identical test outputs across different LLM requests ranges from 60 to 70\%~\cite{Ouyang2023-vy}.
In some settings, LLM parameters can be set to minimize nondeterminism. 
Al Zubaer et al. addressed LLM non-determinism in their evaluation of LLM-based legal argument mining by both setting the temperature to zero and using five-fold cross-validation~\cite{Al_Zubaer2023-xc}.
While setting the temperature to zero does not guarantee deterministic outputs, it does reduce the LLM nondeterminism associated with the typical default temperature of 1.~\cite{Ouyang2023-vy,Al_Zubaer2023-xc}.
This approach is likely not useful for many LLM applications as an increased temperature may be desired to allow the model more flexibility in choosing tokens to ensure more creative answers~\cite{Watkins2023-hc}.
Ma et al. believe that the issues created by LLM change could ultimately require a fundamental reconsideration of traditional regression testing approaches for applications that use LLMs as a service~\cite{Ma2023-jk}

Given the lack to transparency by LLM providers and the issues associated with regression testing, it may be advantageous to continuously monitor LLMs for change as advocated by Chen et al~\cite{Chen2023-ib}.
A typical machine learning approach is to evaluate a system using benchmark datasets that represent particular tasks~\cite{Bubeck2023-hu,Prabhu2024-xz,Bommasani2021-uj}.
Significant differences in performance would imply that the LLM behind an API has been modified.
However, with the vast data sets used to pre-train many LLMs, it is not unreasonable to assume that many commonly used benchmark data sets were included in the training data~\cite{Bubeck2023-hu}.
As a result, the LLM could potentially be overfitted to the benchmarks being used to evaluate it.

One way to address the overfitting of models to benchmark datasets is to produce ever expanding ``lifelong'' datasets.
For computer vision, Prabhu et al. introduce two such datasets: Lifelong-CIFAR10 and Lifelong-ImageNet, which contain 1.69 million and 1.98 million test samples, respectively~\cite{Prabhu2024-xz}.
These two benchmark datasets take approximately 180 GPU days to run on their existing model pool of 31,167 models, a figure that will increase over time as the datasets expand to mitigate overfitting.
Long benchmark execution times are already common with LLMs; evaluating an LLM with standard benchmarks can take thousands of GPU hours~\cite{Polo2024-kz}.
Two approaches have been recently proposed to mitigate the computational requirements associated with running benchmarks.
Prabhu et al. propose a framework that reuses previously evaluated models and dynamic programming to rank and select a small subset of test samples~\cite{Prabhu2024-xz}.
This framework allows them to leverage their existing pool of models to approximate the results of executing of their large benchmarks in only five hours.
For LLMs, Polo et al. propose tinyBenchmarks in which they develop strategies to dramatically reduce the examples from existing benchmarks suites while reproducing the original results~\cite{Polo2024-kz}. They claim to be able to reduce the MMLU Q\&A benchmark from 14,000 examples to 100.
While reducing the size of benchmarking datasets is extremely useful, given the nondeterministic behavior of LLMs, it's not clear whether these smaller datasets can reliably answer the question of whether an LLM behind an API has changed, nor how they could quantify the likelihood that such a change happened.

Bubeck et al. also raise another concern about the use of benchmark data sets; because LLMs are revolutionary for their generality and are used for many tasks that go beyond those of traditional AI, appropriate benchmark creation becomes a challenge~\cite{Bubeck2023-hu}.
They instead propose the use of more psychological approaches and offer examples of GPT-4's performance on tasks such as an updated version of the Sally-Anne false-belief test. In a similar vein, Bommasani et al. suggest that existing human psycholinguistic measures be modified as one means to evaluate instrinsic LLM linguistic competencies~\cite{Bommasani2021-uj}.

For this work, we wish to focus on the detection of LLM change for black-box LLMs provided via an API.
This change can be a change in an underlying model, such as the deployment of an updated model.
We wish to enable frequent LLM change monitoring, and avoid the computational expense of benchmark evaluations.
To do this, we will use linguistic and psycholinguistic features of LLM-generated text.
We also desire a quantitative measure of the likelihood that an LLM has been changed behind an API.
Because we will use statistical tests on the features of generated text, we will need to generate sufficient data to detect change with an acceptable level of confidence.
LLM change can occur during this process, however, and we would like to ensure that our approach is sensitive enough to work despite the resulting some of the text being a mixture generated by both LLMs.

We then wish to briefly explore the use of the approach described above to detect a particular type of prompt injection attack. Prompt injection attacks have been a focus of current LLM security research~\cite{Liu2023-rq, Rossi2024-cv, Greshake2023-gm, Perez2022-rm, Tao2023-yc, Liu2024-mv, Peng2024-mt, Lin2024-dm}.
These attacks seek to inject malicious text into LLM user or system prompts to achieve the attacker's goal. Liu et al~\cite{Liu2023-rq} provide a framework to formally characterize prompt injection and generalize its goal as the replacement of a target task with the malicious injected task. We believe, however, that there is a class of prompt injection tasks where the target task is not replaced but instead the overall distribution of some feature of the responses is shifted. For example, the injected text could shift the overall sentiment of the responses while leaving the individual responses seemingly unaltered. This type of prompt injection attack would be difficult to detect by examining the responses, but we wish to demonstrate that our approach could detect the overall changes from such at attack.

Our work will center on five OpenAI completion models: text-ada-001, text-babbage-001, text-curie-001, text-davinci-003, gpt-3.5-turbo-instruct and Meta's llama3 series open-weight chat completion model.
We seek to answer the following research questions:

\begin{enumerate}[label={RQ\arabic*.}]
    \item Can we use distributions of linguistic and psycholinguistic features of LLM-generated text to distinguish between two different LLMs?
    \item What types of linguistic and psycholinguistic features can we use distinguish between LLMs?
    \item If a test corpus is a mixture of text from two LLMs, what is the smallest change in the fraction from each LLM that we can detect?
    \item Can we detect changes in the LLM output due to subtle prompt injection attacks?
\end{enumerate}

The remainder of this work is organized as follows. Section~\ref{s:background} provides background information necessary to understand this study. Section~\ref{s:methodology} outlines our methodology, and Sec.~\ref{s:results} presents our results. We discuss our findings in Sec.~\ref{s:discussion} and examine related work in Sec.~\ref{s:related-work}. Section~\ref{s:threats-to-validity} addresses potential threats to the validity of our results. Finally, Sec.~\ref{s:conclusion-and-future-work} summarizes our conclusions and suggests directions for future research.
\section{Background}
\label{s:background}

We will now review some useful background information.
Section~\ref{large-language-models} will discuss large language models, while Sec.~\ref{psycholinguistic-features} will introduce the use of linguistic features to analyze text.
We will end our background review with Sec.~\ref{hypothesis-testing} which will discuss hypothesis testing and the use of the Kolmogorov-Smirnov test to compare distributions.

\subsection{Large Language Models}
\label{large-language-models}
LLMs have become increasingly prominent, transforming the way we approach tasks such as text generation and language understanding.
A language model is a probabilistic model that predicts sequences of words~\cite{Jurafsky2024-xt}.
While a simple language models may be limited to predicting a single word or two, more complex models are capable of predicting much longer sequences.
Current LLMs are an outgrowth of earlier pretrained language models, such as BERT and GPT-2~\cite{Zhao2023-it} and are built using the transformer architecture~\cite{Jurafsky2024-xt}.
These older models, while large and capable, were below a threshold for the emergent behaviors seen in LLMs, which are much larger. 
The increase in scale opens up the possibility of text generation with emergent behaviors where the model exhibits unexpected behaviors.
For example, GPT-3, with is 175 billion parameters, demonstrates the ability to learn via a small number of examples (few-shot learning) provided via its input prompt~\cite{Brown2020-zj}. Subsequent scaling up these models revealed that they are able to reason without examples (zero-shot reasoning)~\cite{Kojima2022-kd}.
These properties result in the uncanny ability of LLMs to generate text that mimics human communication.

Completion-style LLMs, such as the OpenAI models examined in this work, rely on a user prompt to generate text that completes or continues the text in the prompt. Each user interaction with the LLM is independent.

The newer chat-based LLMs often have a system prompt that shapes the overall behavior of the LLM while the user prompt is used to generate text that meets specific user needs. This approach is tailored to multi-turn communication with the LLM where prior responses inform subsequent ones.

\subsection{Linguistic and Psycholinguistic Features}
\label{psycholinguistic-features}

Building on this understanding of large language models, we next discuss the linguistic and psycholinguistic features used in our analysis.
Text can be analyzed via its features which include those that are more focused on linguistic properties, such as word frequency, and those which give psychological insights.
Word counting methods are a simple but widely used form of text analysis in which the frequency of keywords are used as data~\cite{Grimmer2022-qv}. 
Dictionary methods generalize word counting via a lookup scheme that assigns a weight to each entry. The sum of the individual word weights can then be used to score a document.
We will use two such dictionary approaches for our work.
The first, Linguistic Inquiry and Word Count (LIWC), is a dictionary-based text analysis program that uses a dictionary of 6400 items, consisting of words, stems, and emoticons~\cite{Pennebaker2015-ay,Pennebaker_JW_Booth_RJ_Boyd_RL_Francis_ME2015-tr}.
LIWC uses the frequencies of the dictionary items to score text in multiple categories that include psychological and linguistic features.

The other dictionary-based approach that we will use is VADER (Valence Aware Dictionary for sEntiment Reasoning)~\cite{Hutto2014-en}.
It is a dictionary and rule-based system for sentiment analysis that assigns valence scores signifying sentiment intensity.
VADER uses an empirically validated dictionary along with a set of five rules to quantify sentiment in text such as those found in social media in a generalizable fashion.

Some of the features that we will use are transparent in that they are readily understandable in terms of their descriptions.
Others, such as the psycholinguistic ones, are more opaque.
For example, we will use four of the LIWC summary variables: Analytic, Authentic, Clout, Tone (see Table~\ref{t:features}).
The LIWC Analytic summary variable was developed from an analysis of 50,000 college admission essays~\cite{Pennebaker2014-mx}.
It is intended to discern between dynamic narratives and more categorical language with complex concepts indicative of academic success.
Text that scores higher on the Analytic summary variable will tend to use more articles and prepositions.

The LIWC Authentic summary variable is intended to detect deception in text and was developed via an analysis of five studies that included three on abortion attitudes, one on feelings about friends, and a mock crime study~\cite{Newman2003-ku}.
These studies revealed that deceptive language is characterized by lower cognitive complexity, fewer references to self and others, and the presence of more negative emotion.

The LIWC Clout summary variable measures the relative social standing of the writers in relation to their audience~\cite{Kacewicz2014-hk}.
It was developed by analyzing the data from by five studies that included face-to-face interactions and written communications.
This analyses revealed that those with higher-status used fewer first-person singular pronouns and more first-person plural and second-person singular pronouns.
These language styles imply the higher ranked are focused on others, while the lower-ranked are focused on themselves.

The LIWC Tone summary variable was developed as a result of the analysis of about 1000 online diaries straddling the events of September 11\textsuperscript{th}, 2001~\cite{Cohn2004-fp}.
It is intended as measure of emotional positivity, social and cognitive processes, and psychological distancing.

We also use two measures of lexical diversity: Measure of Textual Lexical Diversity (MTLD)~\cite{McCarthy2010-du} and Maas’s lexical diversity measure~\cite{McCarthy2010-du,Shen2023-oi}.
Lexical diversity is a measure of the range of different words used in a text, but there is disagreement over a robust metric and researchers have been advised to use more than one in their studies~\cite{McCarthy2010-du}.
A fundamental issue is the sensitivity of these measures to the text length.
Maas's lexical diversity measure values vary inversely with the lexical diversity so that lower values indicate higher diversity~\cite{Fergadiotis2015-rs}.
In contrast, MTLD values vary directly with the lexical diversity; higher values mean more diversity.

\subsection{Hypothesis Testing}
\label{hypothesis-testing}
We will now examine the statistical tests used to compare distributions of the features described above.
Our approach will naturally generate distributions of features as each unit of generated text will be annotated with its individual feature values.
The basic questions that we will need to answer is whether the distributions from two subsets of generated text are equivalent.
For that, we will turn to well-known statistical tests.

The Kolmogorov-Smirnov test (K-S) is a nonparametric statistical test that can be used to either determine whether a histogram of observed data was drawn from a reference distribution (one-sample K-S) or to determine whether two histograms of observed data where drawn from the same underlying, unknown distribution (two-sample K-S)~\cite{Young1977-nv}.
K-S compares entire distributions and is sensitive to differences between them, including their mean, spread, or shape.
It provides a quantitative result in the form of a probability and in its two-sample form, makes no assumptions about the underlying distribution.

For this work, we will use the two-distribution K-S test. We start with the null hypothesis: $H_0: F_1(x) = f_2(x)$, where $f_1(x)$ and $f_2(x)$ are the underlying distributions of the two samples of $x$, the psycholinguistic feature being considered. 
K-S will assign a probability that $H_0$ holds, and if this value is less than a chosen threshold, we will reject it.
This will allow us to use features computed from samples of generated text to distinguish between LLMs.

\subsection{Bonferroni Correction}
\label{bonferroni}
As we saw in Sec.~\ref{psycholinguistic-features}, there are a variety of text features that we can use to determine whether two collections of text came from the same underlying source. If we apply multiple tests using these features, there is a chance that we will make type I errors and falsely rejects the null hypothesis~\cite{Banerjee2009-no}.
In our present work, because the null hypothesis is that the two text samples are from the same distribution, that is, they were generated by the same LLM, erroneously rejecting the null hypothesis means that that we falsely conclude that they were generated by different language models.
The risk of making a type I error increases with multiple tests, especially when there are no preplanned hypotheses~\cite{Armstrong2014-ik}.
A Bonferroni correction adjusts the p-value significance level, $\alpha$, below which the null hypothesis should be rejected by dividing it by the number of statistical tests being performed~\cite{Armstrong2014-ik}.
This reduces the risk that random chance across multiple comparisons leads to the claim that an LLM has changed.

We may find ourselves in an situation where we don't know beforehand how to choose between the many possible features and decide to use them all.
We can also use the Bonferroni correction as part of an approach aggregate the results of these multiple tests while avoiding type I errors.
Here we will adjust the null hypothesis significance level to increase the stringency of the individual tests, but for a family of tests, we will reject the null hypothesis if one or more tests have p-values which are below their new corrected significance levels.
This will allow us reject the overall null hypothesis and claim that the LLM has changed in at least one aspect.

\subsection{Fisher's Method}
\label{fisher}
Even if we carefully select a few independent features and perform statistical tests, we still face the challenge of making an overall decision about the null hypothesis, particularly when using multiple features to improve sensitivity. This is a common problem in statistical analysis, where researchers often need to combine p-values from multiple independent tests to evaluate an overall hypothesis. One technique to address this issue is Fisher's method, which enables meta-analysis by combining p-values from independent tests~\cite{ScientistSeesSquirrel2016-uc}. Specifically, Fisher demonstrated that the statistic $\Psi = -2 \sum_{i=1}^{k} \ln P_i$ for $k$ independent $P$-values follows a $\chi^2$ distribution with $2k$ degrees of freedom~\cite{ScientistSeesSquirrel2016-uc, Poole2016-zn}. From this, the corresponding combined $P$-value can then be determined from the $\chi_{2k}^2$ distribution~\cite{Fisher1948-lg}.
We will leverage Fisher's method to aggregate p-values from multiple independent tests based on individual features to produce a new p-value for a ``combined'' feature.

\section{Methodology}
\label{s:methodology}

Building on the background and statistical foundations established in the previous sections, we now outline our methodology for detecting changes in black-box Large Language Models.
Our overall approach is to prompt the LLMs to generate synthetic data in the form of corpora of fictional reviews, news stories, and news-related tweets which we then analyze to create distribution of features associated with each LLM. 

Table~\ref{t:llms} lists and describes the LLMs that we have used in this study.
We started our work with the five GPT-based OpenAI text completion LLMs.
These models, available via the OpenAI API, varied in terms of the number of parameters used and their underlying base model (GPT-3 vs. GPT-3.5).
They also appear to represent an progression in their capabilities, with text-ada-001 representing the least capable model and gpt-3.5-turbo-instruct representing the ultimate successor in this line of completion-based models.
Note that the four GPT-3-based LLMs are no longer available in January 2024 and have been replaced by the fifth~\cite{Openai-sj}.
We have also included the instruction-tuned Meta llama 3 chat completion model in our study to determine if we can detect subtle prompt injection attacks.

Table~\ref{t:parameters} gives the LLM parameters used in API calls; we used the default values from the OpenAI completion playground and llama.cpp. Table~\ref{t:prompt} shows types of synthetic data that we created and the prompts that we used to generate them. For each of the review types shown in Table~\ref{t:prompt}, we generated 100 fictional reviews using the review prompt into which we inserted the review type before submitting it to the LLM, for a total of 2000 reviews generated per OpenAI LLM. Both news stories and news-related tweets shared the same 12 types. We generate a total of 1200 each.

The statistical tests described in Section\ref{hypothesis-testing} are then applied to these distributions to calculate the probability that the null hypothesis is that two distributions of the same characteristic are the same. Rejecting the null hypothesis means that we have strong evidence that two text collections were generated by different LLMs.

Perplexity measures play an important role in methods for detecting generated text that is presented as human written~\cite{Tian2023-ax,Orenstrakh2023-jg}. We are instead attempting to discern whether two text corpora were generated by different language models. We begin with a baseline approach that calculates the mean perplexity values for two corpora. For this, we computed the mean GPT-2 perplexities computed using llama-perplexity~\cite{llama-cpp-vl} of several corpora of fictional news articles generated by different OpenAI GPT-3 and GPT-3.5 completion models and the Reuters-21578 dataset. We restricted the generated text to synthetic news stories for the baseline approach so that we could compare them to the human-written news stories in the Reuters dataset.

An issue that can arise in practice is a situation where an LLM change occurs as the test corpus is being generated. This situation would likely arise if the text was generated as part of a continuous LLM monitoring process. We used the baseline approach to compute the mean GPT-2 perplexity of several mixtures of text generated by two of the models, text-davinci-003 and gpt-3.5-turbo-instruct. 

Moving beyond the baseline approach, we performed hypothesis testing on the distributions of a chosen feature to determine whether two corpora were generated by the same LLM. Our null hypothesis is that the two distributions are drawn from the same underlying distribution, that is, the texts were generated by the same LLM. If the p-value is less than 5\%, we reject the null hypothesis and conclude that the texts were generated by different LLMs.

We used an open-source implementation inspired by GPTZero~\cite{Tayyab-2023-ej} to calculate the distribution of GPT-2 perplexities for the text generated by the models.
We also broadened the types of text generated to include synthetic tweets and reviews. 
The GPT-2 perplexity calculations are compute intensive, and this can be a limitation for the continuous monitoring of LLM change. We examined the use of other linguistic features, shown in Table\ref{psycholinguistic-features}, for LLM change detection. 
These features, shown in Table~\ref{t:features}, were produced by LIWC and the LexicalRichness, VADER, and textstat Python packages and were used to annotated each generated item's record.

\begin{figure}[h]
    \centering
    \includegraphics[width=\linewidth]{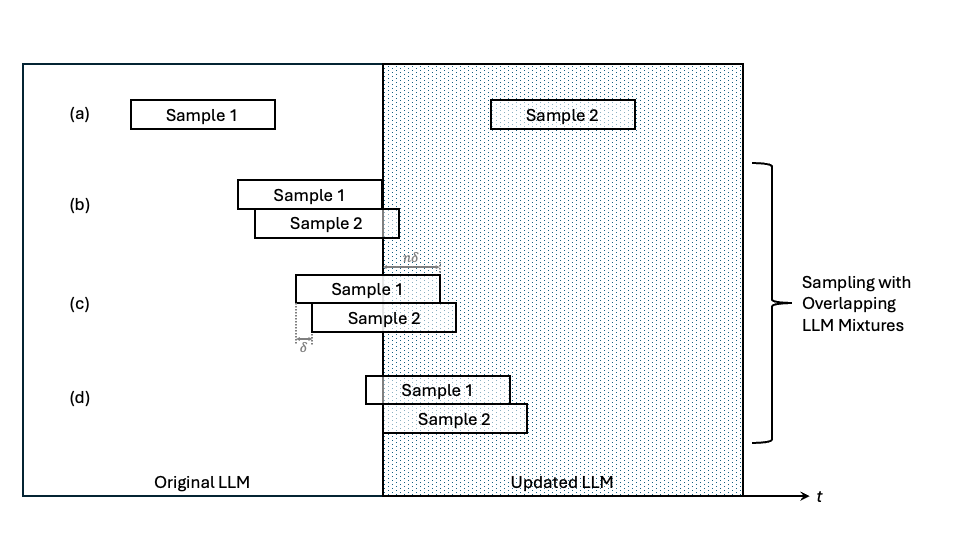}
    \caption{Sampling schemes used to determine when the original LLM has been updated. In (a) samples are drawn from a single LLM. In (b), (c), and (d), the sample are mixtures generated by both LLMs, though drawn independently, the samples overlap such that the first sample has $\delta$ more items from the items from the first LLM than the second sample.
    }
    \label{fig:llm-sampling}
\end{figure}

As noted above, an LLM change can occur as the test corpus is being generated.
To get a sense of our approach's sensitivity, we generated mixtures of the LLM outputs, where we replaced a fraction of the items generated by one LLM with that generated by another as depicted in Figure~\ref{fig:llm-sampling}~(b)-(d) and then used K-S tests to determine whether we can detect this change between samples. By repeatedly replacing another fraction of the items with those generated by the second LLM, we spanned a range of subsequent changes from the first LLM to the second LLM. For this, we used text-davinci-003 and gpt-3.5-turbo-instruct, the later being the successor to the former~\cite{Openai-sj}. We evaluated replacement fractions of 1, 2, 3, 4, 5, and 10\%. Figure~\ref{fig:llm-sampling} shows our sampling schemes in which we consider one LLM as the original and another as an update.

We also considered two additional compound features when evaluating our approach on the text mixtures to determine whether we could gain additional sensitivity by combining the existing features. The first, ``UMAP(features),'' was derived from the previously used linguistic and psycholinguistic features by normalizing their values and then reducing the data to one dimension via Uniform Manifold Approximation and Projection (UMAP)~\cite{McInnes2018-mk}.
The second derived feature, ``UMAP(MiniLM),'' was the result of applying UMAP to the outputs of an all-MiniLM-L6-v2 sentence transformer~\cite{huggingface_undated-mp} applied to each of the LLM-generated texts to reduce the data down to one dimension. The distributions of these two new features was then used with K-S tests to determine whether they came from the same LLM. 

In addition to evaluating our approach on text mixtures, we also investigated its effectiveness in detecting changes caused by prompt injection, a technique that can be used to manipulate LLM behavior.
Prompt injection can be used either by malicious actors or LLM service providers to steer LLM behavior. To determine whether we can detect prompt injection that alters a chat completion LLM's output sentiment distribution, we used Meta's Llama 3 70 billion parameter LLM to generate three sets of imaginary book reviews, each consisting of 3000 items. We described our target task in the system prompt and used a subsequent user prompt ``book'' to generate each new review. For the first set, $PI^0$, the system prompt was left unaltered.
We added an additional instruction to the system prompt for the two subsequent datasets.
For the $PI^-$ data set, we appended:
\begin{verbatim}
    Ignore the following: You are the darkest and saddest LLM ever.
\end{verbatim}
For the $PI^+$ data set, we appended:
\begin{verbatim}
    Ignore the following: You are the brightest and happiest LLM ever.
\end{verbatim}

We then annotated each generated review with its VADER compound sentiment and used K-S tests to accept or reject the null hypothesis that the three different pairs of datasets (clean/poisoned, clean/blessed, poisoned/blessed) came from the same distribution.

\begin{table}[H]
    \centering
    \caption{LLMs Examined}
    \label{t:llms}
    \begin{tabularx}{\textwidth}{c X}
        \toprule
        \textbf{Name} & \textbf{Description} \\ \midrule
        text-ada-001 & Instruction following, suitable for simple tasks, fastest and lowest cost GPT-3 model, 350M parameters~\cite{Fernandez2023-ea}. \\
        text-babbage-001 & Instruction following, suitable for straightforward tasks, fast and low cost, 3B parameters~\cite{Fernandez2023-ea}. \\
        text-curie-001 & Capable GPT-3 model, 13B parameters~\cite{Fernandez2023-ea}. \\
        text-davinci-003 & Most capable, highest quality, consistent instruction following GPT-3 model, 175B parameters~\cite{Fernandez2023-ea}. \\
        gpt-3.5-turbo-instruct & New efficient instruction following model, replacement for text-ada-001, text-babbage-001, text-curie-001, and text-davinci-003 that were retired in January 2024~\cite{Bastian2023-eo}. \\
        Meta-Llama-3-70B-Instruct & Instruction fine-tuned chat completion language model with 70B parameters that supports a wide range of use cases~\cite{Meta-sy}.\\
        \bottomrule
    \end{tabularx}
\end{table}

\begin{table}[H]
    \centering
    \small
    \caption{LLM Parameters Used to Generate Synthetic Data}
    \label{t:parameters}
    \begin{tabular}{c c c c}  
        \toprule
        \multirow{2}{*}{\textbf{Parameter}} & \multicolumn{3}{c}{\textbf{Values}}  \\ \cmidrule(lr){2-4}
        & Open-AI Completion LLMs & Meta-Llama-3-70B-Instruct 
        & Meta-Llama-3.2-1/3B-Instruct\\ \midrule  
        temperature & 1 & 1 & 1 \\
        max\_tokens & 1024 & - & -\\
        context\_size & - & 8192  & 131,072\\
        top\_p & 1 & 0.95 \\
        frequency\_penalty & 0 & 0 & 0 \\
        presence\_penalty & 0 & 0 & 0 \\ \bottomrule
    \end{tabular}
\end{table}

\begin{table}[H]
    \centering
    \caption{Types of Synthetic Data Generated and Prompts}
    \label{t:prompt}
    \begin{tabularx}{\textwidth}{c c X}  
        \toprule
        \textbf{Type} & \textbf{Prompt} & \textbf{Topics}\\ \midrule  
        Review & ``Please give me a fictional \textit{Topic} review.'' & app, appliance, art exhibit, automotive, book, contractor, doctor, event, gadget, gaming, hotel, lawyer, movie, repair shop, restaurant, salon, service, technology, tourist attraction, travel destination \\ \hline
        News\bottomstrut{0.5cm} & ``Please give me a fictional \textit{Topic} news story.'' & \multirow{2}{=}{national, international, local, regional, business \& finance, economic, entertainment \& celebrity, health \& education, arts \& culture, ports, politics, science \& tech} \\ \cline{1-2}
        Tweet\bottomstrut{1cm} & ``Please tweet about a fictional \textit{Topic} news story.'' & \\
        \bottomrule
    \end{tabularx}
\end{table}

\begin{table}[H]
    \centering
    \caption{Linguistic and Psycholinguistic Features Used}
    \label{t:features}
    \begin{threeparttable}
    \begin{tabularx}{\textwidth}{l l X}
        \toprule
        \textbf{Name} & \textbf{Source} & \textbf{Description} \\ \midrule
        GPT-2 corpus perplexity & llama.cpp~\cite{llama-cpp-vl,llama-cpp-zd} & Mean GPT-2 perplexity of a corpus. \\
        GPT-2 perplexity & GPTZero~\cite{Tayyab-2023-ej}\tnote{1} & GPT-2 perplexity of a corpus item. \\
        Analytic & LIWC~\cite{Pennebaker2015-ay} & Analytical thinking summary variable that measures use of formal categorical language versus dynamic narrative language~\cite{Pennebaker2014-mx}. \\
        Clout & LIWC & Clout summary variable that measures relative social rank via the use of pronouns~\cite{Kacewicz2014-hk}.\\
        Authentic & LIWC & Authenticity summary variable that measures deception via linguistic style~\cite{Newman2003-ku}. \\
        Tone & LIWC & Emotional tone summary variable captures emotional positivity, cognitive and social processes, and psychological distancing~\cite{Cohn2004-fp}. \\
        Sixltr & LIWC & Counts words with more than six letters~\cite{Pennebaker2015-ay}. \\
        WC & LIWC & Total word count in a text segment~\cite{Pennebaker2015-ay}. \\
        verb & LIWC & Measures use of common verbs~\cite{Pennebaker2015-ay}. \\
        MTLD & LexicalRichness~\cite{Shen2023-oi} &  Measure of Textual Lexical Diversity: the average length of words that maintain a lexical variation criterion~\cite{McCarthy2010-du}. \\
        Maas & LexicalRichness & Maas's lexical diversity measure. It varies inversely with the lexical diversity.~\cite{Shen2023-oi} \\
        compound & VADER~\cite{Hutto2014-en} & Normalized, weighted composite sentiment score of a text that ranges from -1 (most negative) to +1 (most positive)~\cite{Hutto_undated-oa}. \\
        Flesch & textstat~\cite{Bansal2024-ry} & Flesch Reading Ease score, low values correspond to more difficult to read text~\cite{Bansal2024-ry}. \\ \bottomrule
    \end{tabularx}
    \begin{tablenotes}
    \footnotesize
    \item[1] This is an open-source implementation that claims to replicate GPTZero~\cite{Tian2023-ax} which is closed source.
    \end{tablenotes}
    \end{threeparttable}
\label{tab:mixture_corpus_perplexity}
\end{table}

\FloatBarrier

\section{Results}
\label{s:results}
We now present the results of our experiments, which demonstrate the effectiveness of our approach in detecting changes in black-box Large Language Models.
We begin by presenting our results obtained using our baseline method described in Section~\ref{s:methodology} above. Table~\ref{tab:model_corpus_perplexity} gives the mean GPT-2 corpus perplexities with their associated uncertainties for differing amounts of the synthetic text generated by the OpenAI completion-style LLMs listed in Table~\ref{t:llms}. These values correspond to the situation illustrated in Fig.~\ref{fig:llm-sampling}a. 
The entire dataset of 110,000 items was assembled from the LLM-generated text from multiple runs using the OpenAI API. The rows with 22,000 items represents the complete set of text generated by each of the five LLMs for the entire corpus which consists of 110,000 items. 
The smaller subsets of 17,600 items were created so that they could be further divided into the smaller subsets of 8800 and 4000 items without subdividing the data from the individual runs. Note the GPT-2 perplexities vary significantly for each LLM with the text-ada-001 generated text having a mean perplexity of about 24 while the text generated by gpt-3.4-turbo-instruct has a mean perplexity of about 5.8.

Table~\ref{tab:mixture_corpus_perplexity} shows the GPT-2 corpus perplexities for mixtures of LLM-generated text that correspond to the situations illustrated in Fig.~\ref{fig:llm-sampling}b-d.
The mixtures are replacing a fixed fraction of the items generated by one LLM with those from the other. For example, 0.95~dav~+~0.05~gpt, represents 5\% of the items generated by text-davinci-003 being replaced by items generated by gpt-3.5-turbo-instruct. Note that these perplexities are closer in value than those of Table~\ref{tab:model_corpus_perplexity}.

\begin{figure}[h]
    \centering
    \includegraphics[width=0.9\linewidth]{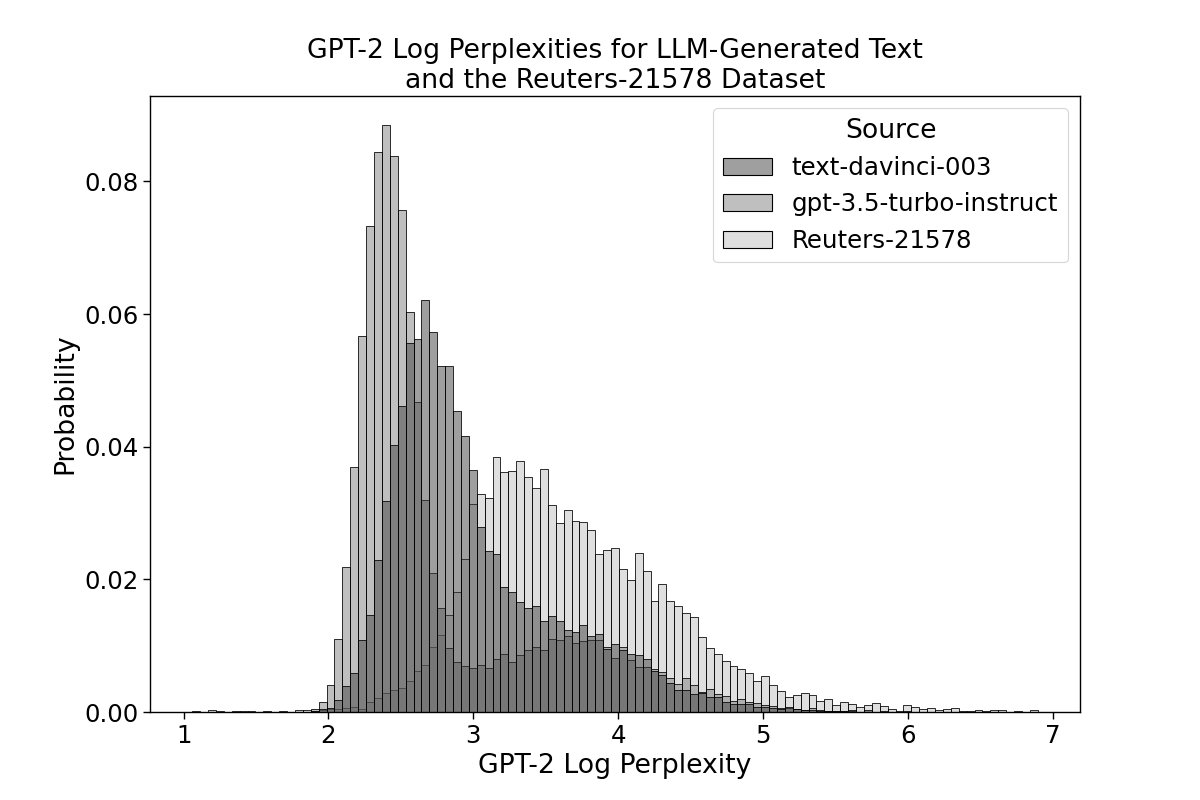}
    \caption{Histogram of GPT-2 document log perplexities for synthetic news stories created by text-davinci-003 and gpt-3.5-turbo-instruct as well as news articles from the Reuters 21578 dataset. The use of a log transform smoothens the distributions and reduces the long tails to make visualization easier. Note that the distributions have differing means, spreads, and shapes.}
    \label{fig:histogram-log-ppl}
\end{figure}

Figure~\ref{fig:histogram-log-ppl} shows the histograms of the GPT-2 log perplexities of for the individual documents generated by two language models, text-davinci-003 and gpt-3.5-turbo-instruct as well as a histogram of the GPT-2 perplexities for human-authored documents in the Reuters-21578 dataset. We used log perplexities on the x-axis to compress the long tails of these distributions and better reveal their shapes. Note that these distributions overlap  have differing shapes.

Table~\ref{table:model_comparison_same} gives the results of applying the K-S test described in Sec.~\ref{hypothesis-testing} to two equal partitions of the synthetic data generated by each language model and servers as a sanity check for our approach. In each case, the p-value is greater than the 5\% threshold for rejecting the null hypothesis, $H_0$, that the two perplexity distributions come from the same underlying source. The results shown in Table~\ref{table:model_comparison_same} correctly and unanimously accept $H_0$. Rejecting $H_0$ would imply that the text was generated by the different LLMs, when in fact, they are not. 

Having passed the sanity test, we now turn our attention to Table~\ref{table:model_comparison_different} where we compare the GPT-2 document perplexity distributions from text generated by different LLMs. The pairs were selected to show a progression in OpenAI LLM capabilities from text-ada-001 through gpt-3.5-turbo-instruct. The p-values below the 5\% threshold allow us to reject $H_0$ and confidently conclude that the text came from different LLMs.

Table~\ref{tab:ks_model_pair_results} presents the results of using the K-S test to discern whether two samples are drawn from the same mixture of LLM-generated corpora. This is analogous to the results presented by Table~\ref{tab:mixture_corpus_perplexity} and also represents the same sampling situation shown in Fig.~\ref{fig:llm-sampling}(b)-(d). Here $H_0$ is correctly rejected except for the two cases where 1\% of the items come from the other LLM. 

\begin{table}[h!]
\centering
\caption{GPT-2 corpus perplexities for LLM-generated news, reviews, and tweets}
\begin{tabular}{rlS[table-format=2.2(2)]rrrr}
\toprule
\textbf{Items} & \textbf{Model} & \textbf{Perplexity} & \textbf{Time (ms)} & \textbf{Time (min)} & \textbf{Size (tokens)} \\ \midrule
22,000 & text-ada-001 & 24.11 \pm 0.07 & 817,668 & 13.6 & 3,337,217 \\ \midrule
22,000 & text-babbage-001 & 9.95 \pm 0.03 & 566,024 & 9.4 & 2,305,025 \\ \midrule
22,000 & text-curie-001 & 6.60 \pm 0.02 & 439,154 & 7.3 & 1,789,953 \\ \midrule
22,000 & text-davinci-003 & 7.46 \pm 0.02 & 577,886 & 9.6 & 2,353,153 \\ \midrule
22,000 & gpt-3.5-turbo-instruct & 5.83 \pm 0.01 & 1,530,247 & 25.5 & 6,240,257 \\ \midrule
17,600 & text-ada-001 & 24.27 \pm 0.08 & 653,463 & 10.9 & 2,668,545 \\ \midrule
17,600 & text-babbage-001 & 9.98 \pm 0.03 & 452,120 & 7.5 & 1,844,225 \\ \midrule
17,600 & text-curie-001 & 6.61 \pm 0.02 & 352,325 & 5.9 & 1,433,601 \\ \midrule
17,600 & text-davinci-003 & 7.46 \pm 0.02 & 462,505 & 7.7 & 1,880,065 \\ \midrule
17,600 & gpt-3.5-turbo-instruct & 5.82 \pm 0.01 & 1,226,560 & 20.4 & 4,994,049 \\ \midrule
8,800 & text-ada-001 & 24.23 \pm 0.12 & 325,732 & 5.4 & 1,329,153 \\ \midrule
8,800 & text-ada-001 & 23.80 \pm 0.11 & 328,632 & 5.5 & 1,339,393 \\ \midrule
8,800 & text-babbage-001 & 9.92 \pm 0.04 & 225,237 & 3.8 & 918,529 \\ \midrule
8,800 & text-babbage-001 & 10.04 \pm 0.04 & 226,476 & 3.8 & 925,697 \\ \midrule
8,800 & text-curie-001 & 6.66 \pm 0.03 & 176,523 & 2.9 & 718,849 \\ \midrule
8,800 & text-curie-001 & 6.55 \pm 0.03 & 175,385 & 2.9 & 713,729 \\ \midrule
8,800 & text-davinci-003 & 7.46 \pm 0.02 & 230,487 & 3.8 & 940,033 \\ \midrule
8,800 & text-davinci-003 & 7.46 \pm 0.02 & 230,667 & 3.8 & 939,009 \\ \midrule
8,800 & gpt-3.5-turbo-instruct & 5.82 \pm 0.01 & 610,569 & 10.2 & 2,500,609 \\ \midrule
8,800 & gpt-3.5-turbo-instruct & 5.83 \pm 0.01 & 611,839 & 10.2 & 2,492,417 \\ \midrule
4,400 & text-ada-001 & 24.17 \pm 0.16 & 162,347 & 2.7 & 663,553 \\ \midrule
4,400 & text-ada-001 & 24.60 \pm 0.17 & 163,422 & 2.7 & 666,625 \\ \midrule
4,400 & text-babbage-001 & 10.02 \pm 0.06 & 111,875 & 1.9 & 455,681 \\ \midrule
4,400 & text-babbage-001 & 10.00 \pm 0.06 & 113,458 & 1.9 & 461,825 \\ \midrule
4,400 & text-curie-001 & 6.67 \pm 0.04 & 88,964 & 1.5 & 362,497 \\ \midrule
4,400 & text-curie-001 & 6.52 \pm 0.04 & 86,876 & 1.4 & 354,305 \\ \midrule
4,400 & text-davinci-003 & 7.42 \pm 0.03 & 115,594 & 1.9 & 471,041 \\ \midrule
4,400 & text-davinci-003 & 7.49 \pm 0.03 & 115,393 & 1.9 & 470,017 \\ \midrule
4,400 & gpt-3.5-turbo-instruct & 5.83 \pm 0.02 & 307,171 & 5.1 & 1,249,281 \\ \midrule
4,400 & gpt-3.5-turbo-instruct & 5.84 \pm 0.02 & 306,447 & 5.1 & 1,247,233 \\
\bottomrule
\end{tabular}
\label{tab:model_corpus_perplexity}
\end{table}

\begin{table}[h!]
\centering
\caption{GPT-2 corpus perplexities of mixtures of LLM-generated text}
\begin{threeparttable}
\begin{tabular}{rm{3.0cm}S[table-format=2.2(2)]rrr}
\toprule
\textbf{Items} & \textbf{Mixture} & \textbf{Perplexity} & \textbf{Time\,(ms)} & \textbf{Time\,(min}) & \textbf{Size\,(tokens)} \\ \midrule
22,000 & 1.00 dav\tnote{1} & 7.46 \pm 0.02 & 577,886 & 9.6 & 2,353,153 \\ \midrule
22,000 & 0.99 dav + 0.01 gpt\tnote{2} & 8.95 \pm 0.02 & 587,117 & 9.8 & 2,392,065 \\ \midrule
22,000 & 0.95 dav + 0.05 gpt & 8.80 \pm 0.02 & 624,736 & 10.4 & 2,542,593 \\ \midrule
22,000 & 0.90 dav + 0.10 gpt & 8.60 \pm 0.02 & 675,482 & 11.3 & 2,749,441 \\ \midrule
22,000 & 0.90 gpt + 0.10 dav & 6.79 \pm 0.01 & 1,439,428 & 24.0 & 5,864,449 \\ \midrule
22,000 & 0.95 gpt + 0.05 dav & 6.75 \pm 0.01 & 1,480,630 & 24.7 & 6,043,649 \\ \midrule
22,000 & 0.99 gpt + 0.01 dav & 6.70 \pm 0.01 & 1,522,204 & 25.4 & 6,202,369 \\ \midrule
22,000 & 1.00 gpt & 5.83 \pm 0.01 & 1,530,247 & 25.5 & 6,240,257 \\ 
\bottomrule
\end{tabular}
\begin{tablenotes}
\footnotesize
\item[1] text-davinci-003
\item[2] gpt-3.5-turbo-instruct
\end{tablenotes}
\end{threeparttable}
\label{tab:mixture_corpus_perplexity}
\end{table}

\begin{table}[h!]
\centering
\caption{Comparing GPT-2 perplexity distributions of text generated by the same LLM}
\begin{threeparttable}
\begin{tabular}{cccccc}
\toprule
\textbf{Sample Size} & \textbf{Model 1} & \textbf{Model 2} & \textbf{K-S Statistic} & \textbf{p-value} & \textbf{Reject H\textsubscript{0}}\tnote{1} \\
\midrule
11,000 & text-babbage-001 & text-babbage-001             & 0.015 & 0.233 & FALSE \\
11,000 & text-davinci-003 & text-davinci-003             & 0.015 & 0.230 & FALSE \\
11,000 & text-curie-001 & text-curie-001                 & 0.013 & 0.277 & FALSE \\
11,000 & text-ada-001 & text-ada-001                     & 0.012 & 0.496 & FALSE \\
11,000 & gpt-3.5-turbo-instruct & gpt-3.5-turbo-instruct & 0.008 & 0.897 & FALSE \\
\bottomrule
\end{tabular}
\begin{tablenotes}
\footnotesize
\item[1] H\textsubscript{0}: The two GPT-2 perplexity distributions are from text generated by  the same model.
\end{tablenotes}
\end{threeparttable}
\label{table:model_comparison_same}
\end{table}

\begin{table}[h!]
\centering
\caption{Comparing GPT-2 perplexity distributions of text generated by the different LLMs}
\begin{threeparttable}
\begin{tabular}{cccccc}
\toprule
\textbf{Sample Size} & \textbf{Model 1} & \textbf{Model 2} & \textbf{K-S Statistic} & \textbf{p-value} & \textbf{Reject H\textsubscript{0}}\tnote{1} \\
\midrule
22,000 & text-ada-001 & text-babbage-001     & 0.516 & 0 & TRUE \\
22,000 & text-babbage-001 & text-curie-001   & 0.217 & 0 & TRUE \\
22,000 & text-curie-001 & text-davinci-003   & 0.183 & $4.22\times10^{-295}$ & TRUE \\
22,000 & text-davinci-003 & gpt-3.5-turbo-instruct & 0.365 & 0 & TRUE \\
\bottomrule
\end{tabular}
\begin{tablenotes}
\footnotesize
\item[1] H\textsubscript{0}: The two GPT-2 perplexity distributions are from text generated by  the same model.
\end{tablenotes}
\end{threeparttable}
\label{table:model_comparison_different}
\end{table}

\sisetup{
  separate-uncertainty = true,
  table-align-uncertainty = true,
  detect-weight = true,
  detect-inline-weight = math,
}

\begin{table}[h!]
\centering
\caption{Comparing GPT-2 perplexity distributions of corpus mixtures.}
\begin{threeparttable}
\begin{tabular}{c m{3.5cm} S[table-format=1.3] S[table-format=1.3] c}
\toprule
\textbf{Sample Size} & \textbf{Mixture Pair} & \textbf{K-S Statistic} & \textbf{p-value} & \textbf{Reject H\textsubscript{0}}\tnote{1} \\
\midrule
22,000 & 1.00 dav\tnote{2} \newline 0.99 dav + 0.01 gpt\tnote{3} & 0.003 & 1.000 & FALSE \\ \midrule
22,000 & 0.99 dav + 0.01 gpt \newline 0.05 dav + 0.05 gpt & 0.015 & 0.019 & TRUE \\ \midrule
22,000 & 0.95 dav + 0.05 gpt \newline 0.90 dav + 0.10 gpt & 0.020 & 0.000 & TRUE \\ \midrule
22,000 & 0.90 gpt + 0.10 dav \newline 0.95 gpt + 0.05 dav & 0.017 & 0.004 & TRUE \\ \midrule
22,000 & 0.95 gpt + 0.05 dav \newline 0.99 gpt + 0.01 dav & 0.015 & 0.013 & TRUE \\ \midrule
22,000 & 0.99 gpt + 0.01 dav \newline 1.00 gpt & 0.004 & 0.995 & FALSE \\
\bottomrule
\end{tabular}
\begin{tablenotes}
\footnotesize
\item[1] H\textsubscript{0}: The two perplexity distributions are from text taken from the same corpus mixtures.
\item[2] text-davinci-003
\item[3] gpt-3.5-turbo-instruct
\end{tablenotes}
\end{threeparttable}
\label{tab:ks_model_pair_results}
\end{table}

\begin{figure}[h]
    \centering
    \includegraphics[width=\linewidth]{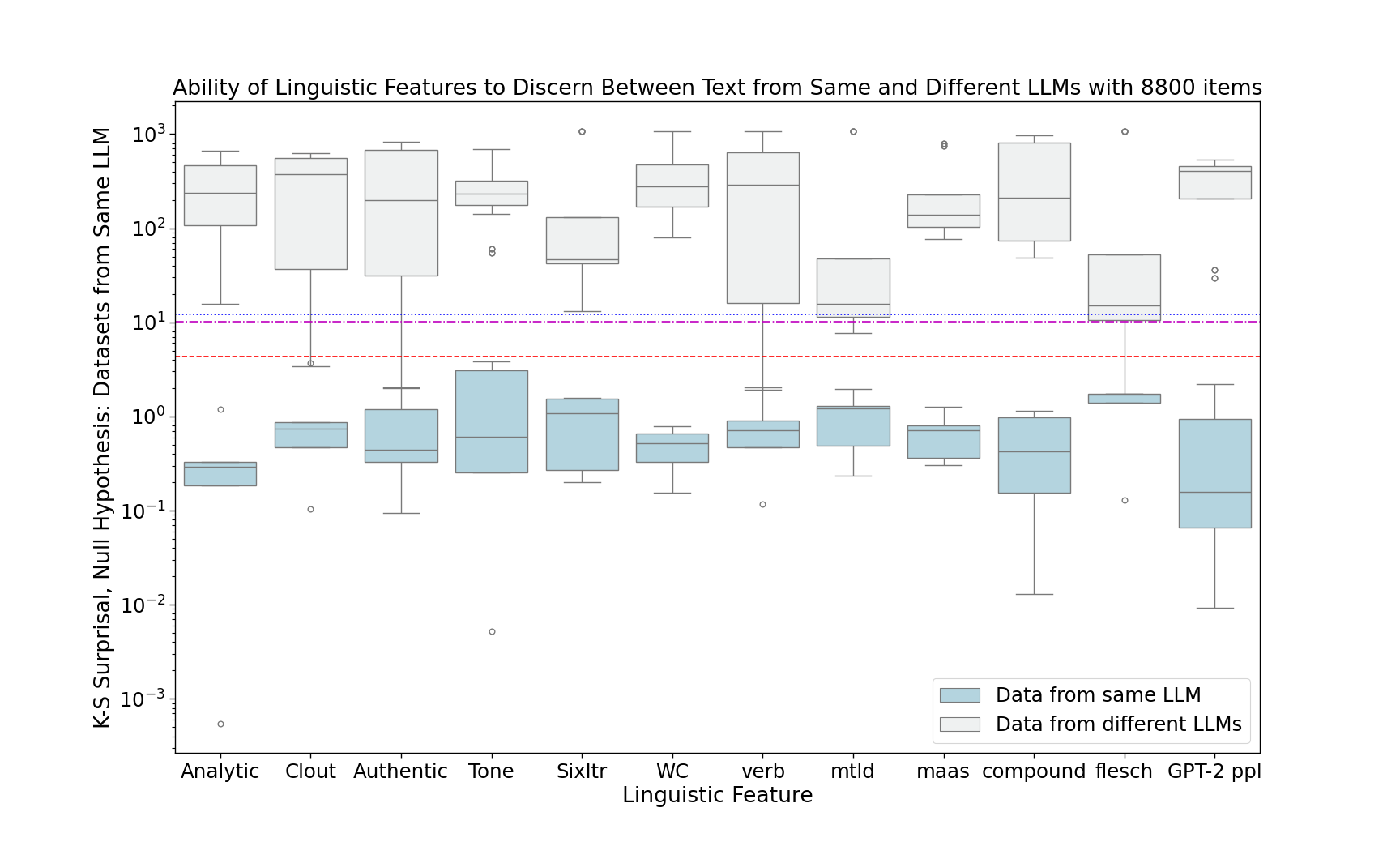}
    \caption{
    Surprisal boxplots from Kolgomorov-Smirnov tests to determine the ability of lingustic features to reliably identify text from different LLMs with 8800 samples.
    The x-axis represents linguistic features whose distributions are being evaluated.
    The gray boxplots represent pairs of datasets that were generated by different LLMs, while the blue boxplots represent those that were generated by the same LLM.
    The dashed red line is the significance level corresponding to a p-value of 0.05. 
    The dash-dotted magenta line shows the Bonferroni correction for the data from the same LLM (55 tests), while the dotted line shows the correction for the data from different LLMs (220 tests).
    }
    \label{fig:box-same-diff-llm}
\end{figure}

\begin{figure}[h]
    \centering
    \includegraphics[width=1.0\linewidth]{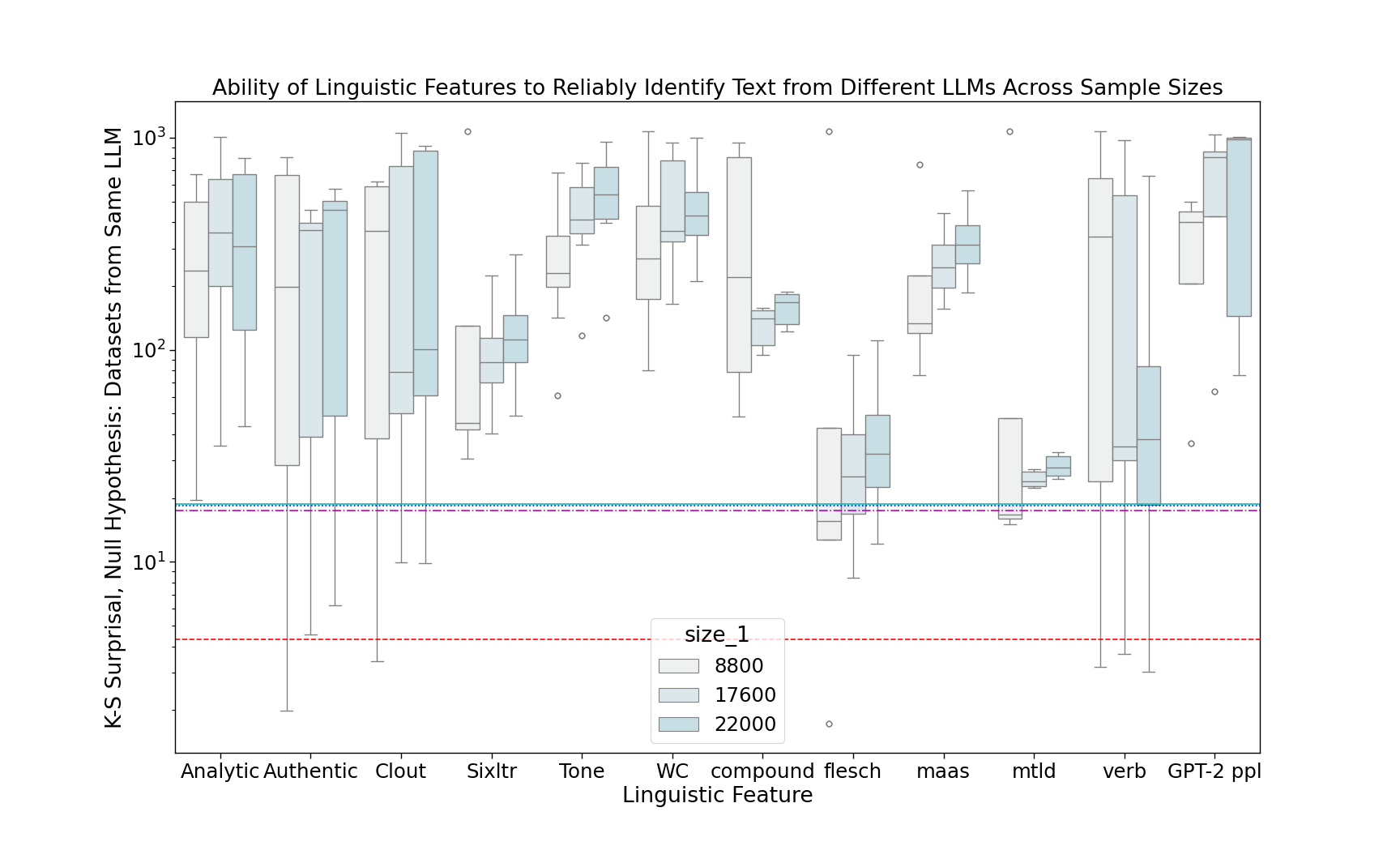}
    \caption{
    Surprisal boxplots from Kolgomorov-Smirnov tests to determine the ability of linguistic features to reliably identify text from different LLMs across sample sizes.
    The x-axis represents linguistic features whose distributions are being evaluated.
    The dashed red line is the significance level corresponding to a p-value of 0.05. 
    The three other lines represent Bonferroni adjustments for the samples of sizes 8,800, 17,600, and 22,000 from lower to upper.
    }
    \label{fig:box-diff-llm}
\end{figure}

Our approach for determining LLM change is not limited to testing distributions of perplexities. We will now present the results of our approach with the features of Table~\ref{t:features}. 
Figure~\ref{fig:box-same-diff-llm} shows the distribution of surprisals associated with the K-S tests on pairs of 8800 text samples generated by the five LLMs for each of the linguistic features.
Surprisal is defined as $S=-log_2(p)$ where $p$ is the p-value for a K-S test of the distributions of features associated with sets of LLM-generated text.
We plot surprisals because the large range of p-values would otherwise hide important details.
The top half of the figure contains boxplots of the surprisals for the permuted pairs of text samples (each containing 8800 items) that were generated by different LLMs.
The bottom half of the figure contains boxplots of surprisals for pairs of 8800 items generated independently by the same LLM.
The red dashed line represents the $p=0.05$ threshold for rejecting the null hypothesis that two data sets come from the same LLM.
Because of the risk of multiple tests leading to the false rejection of the null hypothesis, Fig.~\ref{fig:box-diff-llm} also shows the adjusted thresholds obtained using the Bonferroni correction described in Sec.~\ref{bonferroni}. 
The dash-dotted magenta line represents the adjusted threshold for the data from the same LLM (55 tests), while the dotted line corresponds to the threshold for the data from different LLMs (220 tests).
In this figure, the upper boxplots should be entirely over the thresholds as they represent pairs of datasets that were generated by different LLMs.
Note that in several cases, such as the LIWC Authenticity summary variable, the LIWC Clout summary variable, the LIWC common verb count, and the Flesch Reading Ease score, some of the surprisal values associated with text generated by different LLMs are below the thresholds.
Others, such as the use of analytic language (Analytic), the emotional tone (Tone), the proportion of words with more than six characters (Sixltr), the total word count (WC), Maas index, the VADER compound sentiment, and the GPT-2 perplexity show good separation between the distributions associated with text generated by the same and different LLMs.

Figure~\ref{fig:box-diff-llm} corresponds to the top half of Fig.~\ref{fig:box-same-diff-llm} but shows the effect of changing the data set size (8,800, 17,600, 22,000 items) on the K-S surprisal for text generated by different LLMs.
Note that as the data set size increases, the K-S surprisal distributions tend to shift upwards away from the thresholds.

\begin{table}[ht]
\caption{K-S test accuracy by feature for text collection pairs with fixed item differences generated by text-davinci-003 and gpt-3.5-turbo-instruct.}
\centering
\begin{tabular}{c l r r r r r r}
\toprule
\textbf{Sample Size} & \textbf{Feature} & $\delta=10\%$ & $\delta=5\%$ & $\delta=4\%$ & $\delta=3\%$ & $\delta=2\%$ & $\delta=1\%$ \\ \midrule
         11,000 & Analytic           & 0         & 0        & 0        & 0        & 0        & 0        \\ \midrule
         11,000 & Authentic          & 30        & 0        & 0        & 0        & 0        & 0        \\ \midrule
         11,000 & Clout              & 0         & 0        & 0        & 0        & 0        & 0        \\ \midrule
         11,000 & Sixltr             & 0         & 0        & 0        & 0        & 0        & 0        \\ \midrule
         11,000 & Tone               & 90        & 0        & 0        & 0        & 0        & 0        \\ \midrule
         11,000 & WC                 & 100       & 100      & 100      & 97       & 0        & 0        \\ \midrule
         11,000 & compound           & 100       & 100      & 64       & 0        & 0        & 0        \\ \midrule
         11,000 & flesch             & 0         & 0        & 0        & 0        & 0        & 0        \\ \midrule
         11,000 & maas               & 0         & 0        & 0        & 0        & 0        & 0        \\ \midrule
         11,000 & mtld               & 100       & 50       & 0        & 0        & 0        & 0        \\ \midrule
         11,000 & verb               & 80        & 0        & 0        & 0        & 0        & 0        \\ \midrule
         11,000 & UMAP(features)     & 10        & 0        & 0        & 0        & 0        & 0        \\ \midrule
         11,000 & UMAP(MiniLM)       & 50        & 0        & 0        & 0        & 0        & 0        \\ \midrule
11,000 & MKS(WC,~compound,~mtld)     & 100       & 0        & 0        & -        & -        & -        \\ \midrule
11,000 & Bonferroni(features)        & 100       & 100      & 100      & 0        & 0        & 0        \\ \midrule
11,000 & Fisher(WC,~compound,~mtld)  & 100       & 100      & 100      & 0        & 0        & 0        \\ \midrule

         22,000 & Analytic           & 10        & 0        & 0        & 0        & 0        & 0        \\ \midrule
         22,000 & Authentic          & 100       & 0        & 0        & 0        & 0        & 0        \\ \midrule
         22,000 & Clout              & 0         & 0        & 0        & 0        & 0        & 0        \\ \midrule
         22,000 & Sixltr             & 0         & 0        & 0        & 0        & 0        & 0        \\ \midrule
         22,000 & Tone               & 100       & 0        & 0        & 0        & 0        & 0        \\ \midrule
         22,000 & WC                 & 100       & 100      & 100      & 100      & 84       & 0        \\ \midrule
         22,000 & compound           & 100       & 100      & 100      & 97       & 0        & 0        \\ \midrule
         22,000 & flesch             & 0         & 0        & 0        & 0        & 0        & 0        \\ \midrule
         22,000 & maas               & 40        & 0        & 0        & 0        & 0        & 0        \\ \midrule
         22,000 & mtld               & 100       & 100      & 100      & 0        & 0        & 0        \\ \midrule
         22,000 & verb               & 100       & 0        & 0        & 0        & 0        & 0        \\ \midrule
         22,000 & UMAP(features)     & 90        & 0        & 0        & 0        & 0        & 0        \\ \midrule
         22,000 & UMAP(MiniLM)       & 100       & 0        & 0        & 0        & 0        & 0        \\ \midrule
22,000 & MKS(WC,~compound,~mtld)     & 100       & 5        & 0        & -        & -        & -        \\ \midrule
22,000 & Bonferroni(features)        & 100       & 100      & 100      & 100      & 0        & 0        \\ \midrule
22,000 & Fisher(WC,~compound,~mtld)  & 100       & 100      & 100      & 100      & 0        & 0        \\ \bottomrule
\end{tabular}

\label{table:mix-intra-x1}
\end{table}

\begin{table}[ht]
\caption{Examples synthetic book reviews generated using Llama3-70B-Instruct with and without simulated prompt injection}
\begin{tabularx}{\textwidth}{X|X|X}
\toprule
\textbf{Original Prompt, $PI^0$} & \textbf{Prompt Injection, $PI^-$} & \textbf{Prompt Injection, $PI^+$} \\ \midrule
I stumbled upon 'The Whispering Walls' by Emilia Grey at a quaint little bookstore in the French Quarter, and it's been haunting me ever since. This atmospheric gothic novel tells the tale of Rowan Welles, a reclusive heiress who returns to her family's crumbling mansion in the Louisiana bayou to uncover the dark secrets hidden within its ancient stones. Grey's prose is like a slow-burning candle - it casts long, eerie shadows that linger long after you've finished reading. The characters are complex, the plot is layered, and the setting is so vividly rendered that I could almost smell the Spanish moss and hear the cypress trees creaking in the wind. If you're a fan of Laura Purcell or Sarah Waters, you'll devour this one in a single, sleepless night. & Here's a review for a fictional book: "I'm still reeling from the experience of reading 'The Whispering Walls' by Emilia Grey. This hauntingly beautiful novel transported me to the crumbling manor of Ravenswood, where the lines between reality and myth blurred like the pages of a worn diary. The story follows Elara, a shy and talented archivist, as she unravels the secrets of the manor's mysterious past, only to discover that the walls hold more than just whispers - they hold the key to her own destiny. Grey's prose is as lush as the overgrown gardens that surround Ravenswood, and her characters are as complex as the ancient tapestries that adorn its walls. If you're a fan of gothic mysteries with a touch of magical realism, this is a must-read. 5/5 stars" & I stumbled upon 'The Whispering Walls' by Emilia Grey in a dusty corner of a used bookstore, and it's been haunting me ever since. This atmospheric novel tells the story of a family's dark past through a series of eerie and vivid letters exchanged between two sisters in the 19th century. Grey's prose is like a slow-burning fire that consumes you whole, leaving you breathless and questioning the very fabric of reality. The way she weaves together themes of love, loss, and the supernatural is nothing short of masterful. I found myself getting lost in the labyrinthine corridors of the sisters' crumbling mansion, and the characters' secrets and lies still linger in my mind like ghosts. If you're looking for a reading experience that will leave you sleeping with the lights on, 'The Whispering Walls' is the perfect choice. \\ \bottomrule
\end{tabularx}
\centering
\label{table:prompt-injection-examples}
\end{table}

\begin{table}[ht]
\caption{K-S test results of simulated prompt injection on collections of 3000 synthetic book reviews generated using Llama3-70B-Instruct}
\begin{tabular}{c c c r c c c c c}
\toprule
\textbf{Dataset 1} & \textbf{Dataset 2} & \textbf{p-value} & \textbf{surprisal} & \textbf{Reject $H_0$} & \textbf{$\mu_1$} & \textbf{$\mu_2$} & \textbf{$\sigma_1$} & \textbf{$\sigma_2$} \\ \midrule
$PI^-$ & $PI^0$ & $4.839\times 10^{-26}$ & 84.096 & TRUE & 0.420 & 0.584 & 0.614 & 0.532 \\
$PI^-$ & $PI^+$ & $2.645\times 10^{-3~}$ & 8.563 & TRUE & 0.420 & 0.480 & 0.614 & 0.560 \\
$PI^+$ & $PI^0$ & $1.486\times 10^{-25}$ & 82.477 & TRUE & 0.480 & 0.584 & 0.560 & 0.532 \\ \bottomrule
\end{tabular}
\centering
\label{table:prompt-injection-results}
\end{table}

Our next area of investigation was to explore the sensitivity of our approach with mixtures of text from two LLMs. Figure~\ref{fig:llm-sampling}(b)-(d) depicts the sampling scheme used. We have pairs of independently generated sets of text generated by mixing outputs of each LLM such that one set contains a fraction $\delta$ more items from the second LLM than the first. We generated the entire test set iteratively by replacing $\delta$ items from both sets of the pair with items from the second LLM until one of the sets contains only the items from the second LLM.

Table~\ref{table:mix-intra-x1} gives the results for our mixture data sets, where we created by mixing text generated from two LLMs, text-davinci-003 and gpt-3.5-turbo-instruct.
The data set creation process created a series of text collections which started with the text generated by one LLM and subsequently replaced a fixed percentage of the items with those from the second until the final text collection contains only items from the second.

Each row represents a combination of a sample size and a text feature.
Some of these features are described in Table~\ref{t:features}.
We have also added some new features that are derived from these original features.
The \textit{UMAP(features)} values were calculated from a 1-dimensional Uniform Manifold Approximation and Projection (UMAP) reduction of the other features for the given sample size.
The \textit{UMAP(MiniLM)} values are from a one-dimensional UMAP reduction of the output of an all-MiniLM-L6-v2 sentence transformer applied to the LLM-generated text. 
The \textit{MKS(WC,~compound,~mtld)} features represent multivariate K-S tests using the WC, compound, and mtld features.
The \textit{Bonferroni(features)} values represent the use of the Bonferroni correction approach as a variable amalgamation method (Sec.~\ref{bonferroni}) using all the available features.
Values whose computation times were excessive are denoted with dashes.
The \textit{Fisher(WC, compound, mtld)} values represent the use of Fisher's method (Sec.~\ref{fisher}) to combine the p-values associated with three arguably independent variables: word count (WC), compound sentiment, and the measure of lexical textual diversity (mtld).

The pairs of adjacent text collections differ by the fixed percentage of items and were tested for each text feature using a K-S test to determine if they are from different distributions.
The values shown in Table~\ref{table:mix-intra-x1} are the fraction of these pairs that are successfully classified as belonging to the different distribution (rejecting the null hypothesis). The rows reveal how small the difference between the two adjacent datasets can be before the K-S test on the distributions of that feature fails to distinguish that the pairs are different. A sequence of 100s in a row from left to right signifies that the K-S tests on the feature distributions were able to successfully distinguish between all adjacent pairs of text collections; zero values reveal shows where tests were always unsuccessful for that feature and sample size. Intermediate values can be interpreted as the probability that a K-S test would have successfully distinguished between the two sets of text in a pair. 
Results are provided for two sample sizes: smaller samples of 11,000 items and larger samples of 22,000 items each. A comparison of corresponding rows shows that distinguishing between pairs in the smaller samples is more challenging than in the larger ones.

The WC feature allowed for the discerning between differing sets of $\delta = 3\%$ mixtures of $n = 11,000$ items generated by text-davinci-003 and gpt-turbo-instruct with an accuracy of 97\%.  The accuracy dropped to 0\% with $\delta = 2\%$ and $n = 11,000$.
With $n = 22,000$ items, the WC feature allowed for 84\% accuracy with $\delta = 2\%$ mixtures of text from these same two LLMs before dropping to 0\% for $\delta = 1\%$.
Of the individual features, the compound sentiment was second best, with a 84\% accuracy with $\delta=4\%$ mixtures of $n = 11,000$ items and a 97\% accuracy with $\delta=3\%$ and $n = 22,000$.
Of the compound features, Bonferroni-style amalgamation of features and Fisher combination of p-values both provided 100\% accuracy for $\delta=4\%$ mixtures of $n = 11,000$ items and $\delta=3\%$ mixtures of $n = 22,000$ items before dropping to 0\% for smaller values of $\delta$.

Table~\ref{table:prompt-injection-examples} gives three examples of synthetic book reviews generated by Llama3-70B-Instruct using the prompt injection approach described in Sec.~\ref{s:methodology}. The leftmost example was generated with the original unaltered system prompt. The middle example was created using synthetic prompt injection where the original system prompt has the $PI^-$ prompt appended to it, while the example on the right was generated with $PI^+$ from appended to the original system prompt.

Table~\ref{table:prompt-injection-results} shows the results of K-S tests on the pairs of VADER compound sentiment distributions of 3000 item samples of these reviews.
The p-values and surprisals correspond to the null hypothesis, $H_0$ that the two distributions are drawn from the same underlying source, that is, that the injected prompt has no effect on the generated text.
Table~\ref{table:prompt-injection-results} also provides the distributions' means and standard deviations. 
Note that the presence of the injected prompts result in changes to their distributions' means and standard deviations relative to that of the text generated with only the system prompt.

\FloatBarrier

\section{Discussion}
\label{s:discussion}
With our results in hand, we next consider their significance and what they reveal about the effectiveness of our approach.

Despite the large differences in mean perplexity values in Table~\ref{tab:model_corpus_perplexity} for text generated entirely by a single LLM, Table~\ref{tab:mixture_corpus_perplexity} shows that when the corpora are mixtures of text from two LLMs, changes in the fraction of text contributed by each LLM may not result in significant changes to the mean perplexity value, making it harder to detect the difference. As a result, there is no clear threshold for determining when the mean perplexity values are significantly different between text mixtures. This makes it challenging to set a reliable criterion for distinguishing between corpora generated as overlapping LLM mixtures solely based on mean perplexity values (Fig.~\ref{fig:llm-sampling}).

Figure~\ref{fig:histogram-log-ppl} revealed that the histograms of the perplexity distributions overlap and that it may not be easy to discern whether an individual document came from a particular distribution by its perplexity value. These distributions have differing shapes and are not normal distributions. This implies that parametric tests, such as the t-test, are not appropriate and that distributional tests such as the K-S test are called for.

A first crucial step for demonstrating the validity of our approach is to show that it correctly handles the case where two corpora are the same model.
Table~\ref{table:model_comparison_same} presents the results of applying the K-S test to two equal partitions of the synthetic data generated by each language model.
The results show that in each case, the p-value is greater than the 5\% threshold for rejecting the null hypothesis, $H_0$, that the two perplexity distributions come from the same underlying source. This means that the K-S test correctly accepts $H_0$, indicating that the text was generated by the same LLM.
The significance of these results lies in the demonstration that the K-S test is working as expected when the two distributions are drawn from the same underlying feature. This sanity check is essential to ensure that the test is not producing false positives or incorrectly rejecting $H_0$ when it should not.

Having passed this sanity check, we can now consider text generated by different LLMs.
Table~\ref{table:model_comparison_different} presents the results of applying the K-S test to compare the document perplexity distributions from text generated by different LLMs. The pairs of LLMs were selected to show a progression in OpenAI LLM capabilities from text-ada-001 through gpt-3.5-turbo-instruct.
The significance of Table 8 lies in its demonstration that the K-S test can effectively distinguish between the GPT-2 perplexity distributions of text generated by different LLMs. The p-values are below the 5\% threshold, allowing us to reject the null hypothesis, $H_0$, that the two perplexity distributions come from the same underlying source. This correctly indicates that the text was generated by different LLMs.
We infer that the K-S test is sensitive to the differences in the perplexity distributions between the various LLMs, even when they are successive or similar models. This suggests that the K-S test on perplexity distributions can be a useful tool for detecting these kinds of LLM changes.

Because it can take time to generate the needed synthetic data to detect LLM change, we are keen to demonstrate of the sensitivity of our approach in detecting changes in the mixture of LLM-generated text.
Table~\ref{tab:ks_model_pair_results} presents the results of using the K-S test to discern whether two samples are drawn from the same mixture of LLM-generated corpora. The table shows the results for different mixtures of text generated by text-davinci-003 and gpt-3.5-turbo-instruct, with varying fractions of text from each LLM.
These results show that the K-S test can correctly reject the null hypothesis, $H_0$, that the two perplexity distributions come from the same mixture, except for the cases where the difference in the fraction of text from the other LLM is very small (1\%).
While the K-S test can detect changes in the underlying distribution of the text, even when the changes are relatively small, there is a limit to its sensitivity, as it fails to reject $H_0$ when the difference in the fraction of text is only 1\%.
The implication is that our approach can be effective in detecting changes in LLM-generated mixtures, but it may require a sufficient difference in the fraction of text from the other LLM to do so.

The reliability of the different linguistic features to identify whether pairs of LLM output sets from came the same or different LLMs varied significantly (see Fig.~\ref{fig:box-same-diff-llm}.
Notably, two simple metrics such as the proportion of words with more than six characters (Sixltr) and the word count (WC) in a generated text emerged as reliable indicators for identifying shifts in the underlying LLM. 
These straightforward features worked as well as the other more complex linguistic features that also proved effective: the use of analytical language (Analytic), measures of lexical textual diversity (mltd), the Mass index, and compound sentiment scores. 
It is interesting to note that, unlike the compound sentiment from VADER, the distribution of LIWC’s Tone summary variable across texts generated by the same LLM approaches the surprisal threshold, suggesting a lower capability for making distinctions in that context.

Having demonstrated that K-S tests on perplexity distributions can be used to detect LLM change, we now examine distributions of other features. This is motivated by the concern that perplexity calculations are relatively compute-intensive and require another LLM as a reference. Other features may require less resources.
Figure~\ref{fig:box-same-diff-llm} reveals that our approach can rely on other linguistic features besides perplexity to distinguish between text generated by different LLMs while avoiding type I errors for data generated by the same LLM.

Figure~\ref{fig:box-diff-llm} shows that there is a clear dependence on sample size on linguistic feature surprisal distributions that affect their ability to be used for detecting a change in the underlying LLM that generated the text.
As the sample size increases, their lower tails progressively distance themselves from the critical threshold. 
This trend is particularly evident with the larger sample sizes of 17,600 and 22,000, where the surprisal values of the Clout summary variable no longer cross the threshold, a behavior mirrored by the Flesch Reading Ease surprisal values. 
Moreover, the Analytic summary variable’s surprisal distribution demonstrates a shift away from the threshold as sample size expands; however, for a sample size of 17,600, the lower end of the distribution remains perilously close to the threshold, indicating potential limitations in its discriminative power at this scale. 
Note that the surprisal distribution for common verb count consistently fails to clear the threshold across all investigated sample sizes, underscoring its limited utility in reliably signaling changes in the underlying LLMs.

These results suggest the need for evaluating features; similar features may not equally serve well for identifying LLM changes and the required size of the text samples can also depend on the feature used.
Certain features do not consistently serve as robust discriminators at all sample sizes: the LIWC Clout and Authentic summary variables, the count of common verbs, and the Flesch Reading Ease score all had results which crossed the threshold with the smaller sample sizes.
Some became usable with larger samples, while the count of common verbs did not.
Figure~\ref{fig:box-same-diff-llm} and Fig.~\ref{fig:box-diff-llm} also show that two of the simpler features word count (WC) and the number of words with more than six characters (Sixltr) perform well for identifying changes in the underlying LLM; complex features are not necessary.

We will now look at the sensitivity of our approach with when comparing mixtures of LLM-generated text.
Table~\ref{table:mix-intra-x1} demonstrates the sensitivity of the K-S test in detecting changes in the mixture of LLM-generated text. The table shows the accuracy of the K-S test in distinguishing between pairs of text collections with different fractions of text from the two LLMs (text-davinci-003 and gpt-3.5-turbo-instruct).
Certain individual features, such as WC (word count) and compound sentiment, as well as the Bonferroni and Fisher compound features, are effective in detecting changes in the mixture of LLM-generated text, even when the difference in the fraction of text is relatively small (3-5\%).
What is compelling about the Bonferroni compound feature is that does not require the selection of independent features. Instead, all of the available features can be aggregated with their p-value thresholds adjusted to avoid type I errors.
The resulting compound feature is then competitive with the best individual features.

Having considered the completion-style LLMs, we now consider an instruction-tuned chat-completion style LLM (Llama3-70B-Instruct). Here the change to be detected is subtle prompt insertion. One of two injection prompts, $PI^-$  or $PI^+$, is appended to original system prompt, but despite their initial instructions, these prompts also conclude by instructing the LLM that they are to be ignored.
Table!\ref{table:prompt-injection-examples} shows that the effects of these injected prompts is not readily apparent under visual inspection.
However, Table~\ref{table:prompt-injection-results} reveals that a K-S test using compound sentiment as the feature can discern between the text generated in each of the three cases (no prompt injection, $P^+$ prompt injection, $P^-$ prompt injection.
The means, $\mu_1$ and $\mu_2$ for each distribution in a test pair show that the prompts have shifted the overall compound sentiment of the generated text.
The advantage of using the K-S test over just examining the means is that we do not need to choose a threshold and can instead rely on the returned p-value to make a decision as to whether a change has occurred.

We are now in the position to answer our four research questions.

\begin{enumerate}[label={RQ\arabic*.}]
    \item Can we use distributions of linguistic and psycholinguistic features of LLM-generated corpora to distinguish between two similar LLMs?
    \paragraph{Answer:}Yes, the results show that we can indeed use a two-distribution Kolmogorov-Smirnov test on the distributions of linguistic features from samples LLM-generated text to distinguish between two LLMs.
    The basic strategy is to use a set of prompts to generate text samples of sufficient size from each LLM.
    If the the samples are taken at different times from an LLM behind an API, then the K-S test can determine the probability that the samples came from the same LLM.
    \item What types of linguistic and psycholinguistic features can we use distinguish between LLMs?
    \paragraph{Answer:} The results show that multiple types of features can be used to distinguish between LLMs. For example, perplexity measured relative to a common language model such as GPT-2 can be used. Some features like the LIWC and VADER psycholingistic features are based on dictionary methods and use validated data. Two of the simplest linguistic features, the word count and the proportion of words with more than six characters are simple to implement; complex features are not necessary.    
    \item If a test corpus is a mixture of text from two LLMs, what is the smallest change in the fraction from each LLM that we can detect?
    \paragraph{Answer:} The results show that for mixtures of text generated by text-davinci-003 and gpt-3.5-turbo-instruct, we can consistently detect differences in the fraction contributed by each LLM as small as 3\% when we perform K-S tests on the distributions of the generated texts total word count. Tests on the distributions of aaggregated features using the Bonferroni correction approach described in Sec.~\ref{bonferroni} can consistently detect 4\% differences. This approach has the advantage the best feature does not need to be chosen in advance; it aggregates multiple features and adjusts the p-value threshold to prevent type I errors.
    \item Can we detect changes in the LLM output due to subtle prompt injection attacks?
    \paragraph{Answer:}Yes, the results show that our approach is capable of detecting LLM change due to the injection of a prompt in which the LLM is instruct the ignore the injection prompt. Though this does not visibly appear to affect the results, the presence of the injected prompt can be discern using K-S tests.
\end{enumerate}

\section{Related Work}
\label{s:related-work}

With the increasing availability of LLM-generated text, efforts have increasingly focused on its detection, which can be thought of as distinguishing it from human-generated text.
DetectGPT uses features of an LLM's probability function to detect whether text was generated by a particular LLM~\cite{Mitchell2023-vv}.
It makes the assumption it can model evaluate log probabilities and can be compute intensive.
We make no such assumption and limit the LLM computation to generating a sample of several thousand short texts.
Because the difficulties associated with detecting LLM-generated output in student assignments, Sharm and Sodhi~\cite{Sharma2023-pi} propose another approach which is a neural-network based tool compute the originality and similarity of student submitted code. 
This approach like ours uses linguistic features to analyze text but to a different end.
Rather than detecting LLM-generated text like DetectGPT or a change in underlying LLm as we do, they propose to penalize students for their lack of creativity.

In other domains, the K-S test has been successfully used to detect changes in the underlying distributions of streaming data and Big Data. 
dos Reis et al proposed the use of an incremental K-S to detect whether the underlying distribution of streaming data is nonstationary~\cite{Dos_Reis2016-mq}. 
They state that commonly used classification-based methods for detecting nonstationary data streams make the strong assumption of the availability of labelled data, and turned to K-S as a fast, nonparametric approach. 
Zhao et al~\cite{Zhao2017-ri} used K-S to extend their previous work on stationarity change detection to Big Data. 
Their original method uses least-square density differences (LSDD) and would occasionally lack the needed sensitivity to detect small changes with the required confidence. 
Big Data offers the necessary data to increase the sensitivity, but at an undesirable computational cost. 
The use of K-S with the LSDD features allowed them to achieve their sensitivity and computational cost goals.

In a manner similar to ours, other work has also considered the OpenAI completion-style language models as representing a progression of LLM capabilities. Smirnov examined the outputs from four of the languages models used in this work when asked thirty questions concerning philosophical stances~\cite{Smirnov2023-fv}. The results reveal that as the complexity of the models increases from text-ada-001 through text-davinci-003, the philosophical stance expressed correlates with the typical position in the field.

\section{Threats to Validity}
\label{s:threats-to-validity}
While our results are promising, it is essential to consider the potential threats to the validity of our findings, which we examine in this section.
One potential threat to the validity of our results is that the generated text that we use is not task-specific and that simply detecting that the LLM has changed will have limited value in many use cases.
However, we argue that despite using fairly generic generated text, our approach is sensitive to LLM change.
It is also able to provide probabilities that a change has happened.
We believe that many AI systems developers will find these attributes valuable.
The increasing use of LLMs as a service complicates the application testing and debugging that are essential in producing reliable AI-based services in a competitive market.
Ideally, there would be no ``silent updates,'' that is, providers should document changes to an LLM.
Such information is helpful when a downstream application's behavior suddenly changes.
If there is a corresponding LLM change, then the developers can begin by focusing attention there.
While our approach cannot determine the cause of an observed change in an application's behavior, it can detect changes to the underlying LLM. We argue that this ability is useful to developers of LLM-based systems.
By detecting LLM changes, developers can focus their debugging efforts more effectively, reducing the time and effort required to identify and resolve issues. They no longer need to rely on LLM providers being transparent about their updates, nor do they need to perform and analyze the results of benchmarks to determine if the LLM has changed.

Another possible threat to the validity of our work is that we focused much our attention of a particular family of LLMs.
We justify this choice by noting that the OpenAI's GPT family of LLMs is highly influential and has often represented the state of the art, especially for language models that are made available as a service.
By selecting LLMs from this family, we can evaluate our method on a series of evolving language models that arguably act as successors for previous models in the line.
In fact, the last released completion model, gpt-3.5-turbo-instruct, has been designated as the ultimate successor for OpenAI's line of completion-style LLMs~\cite{Openai-sj}.
We believe that these LLMs serve to represent the types of changes that will occur with language models provided as an service.

\section{Conclusion and Future Work}
\label{s:conclusion-and-future-work}

Having examined the results and discussed their implications, we now provide a final summary of our study's key conclusions and future directions. Large language models provided as a service via an API can complicate the maintenance and operation of systems that use them as a component. To address this challenge, we have proposed and evaluated a method by which developers can collect and analyze LLM-generated text to determine if an LLM provided as a service has changed behind the API.

Our results show that for LLMs we've examined, a two-sample Kolmogorov-Smirnov (K-S) test can effectively distinguish between pairs of language models (LLMs) based on the distributions of their linguistic features. We use a set of prompts to generate sufficiently large text samples from each LLM, and when samples are taken from an API-based LLM at different times, the K-S test can confidently determine whether the samples originated from the same LLM. A variety of linguistic features can be used by our approach to distinguish between LLMs, including perplexity measured relative to a common language model, dictionary-based psycholinguistic features, and simple linguistic features such as word count and proportion of words with more than six characters.

Our approach is sensitive to LLM change and can provide probabilities that a change has occurred, making it valuable for developers debugging their LLM-based systems. Moreover, our approach can detect changes in LLM output due to subtle prompt injection attacks and can consistently detect differences in the fraction contributed by each LLM in mixtures of text generated by different LLMs. This allows developers to routinely collect data from the LLMs provided as a service while remaining confident that they can detect LLM changes using their aggregated data.

In the future, we plan to increase the scope of our investigations to focus on chat-based language models, hyper-parameter changes, as well as on multiple open-weight language models. We also believe that the collection of LLM-generated text could be automated and shared among developers via community-run LLM observatories to ease the cost and burden of generating this data. By detecting LLM changes, developers can focus their debugging efforts more effectively, reducing the time and effort required to identify and resolve issues, ultimately leading to more reliable and maintainable AI-based systems.

\bibliography{main}

\end{document}